# Real-time low-resource phoneme recognition on edge devices

**Yonatan Alon**

Independent researcher

yonatankarimish@gmail.com

Abstract:

While speech recognition has seen a surge in interest and research over the last decade, most machine learning models for speech recognition either require large training datasets or lots of storage and memory. Combined with the prominence of English as the number one language in which audio data is available, this means most other languages currently lack good speech recognition models.

The method presented in this paper shows how to create and train models for speech recognition in any language which are not only highly accurate, but also require very little storage, memory and training data when compared with traditional models. This allows training models to recognize any language and deploying them on edge devices such as mobile phones or car displays for fast real-time speech recognition.

Introduction:

It is estimated that as many as 86.8% of the people in the world do not speak English at all (Crystal, 2003). However, since the advent of computers nearly the entire computer ecosystem has been written in English. When it comes to making this ecosystem accessible to end users in products, services and utilities, it becomes necessary to translate it to a multitude of other languages.

Services using machine learning algorithms are no different than the rest. However, when translating machine learning based systems for end users it is often not enough to just translate the user interfaces. When these systems have an audio related component, this could mean translating the entire data pipeline. Such a process involves recording potentially massive amounts of speech in the target language and training new models using the newly collected speech data.

Collecting such amounts of data and training models to process it are two major barriers faced by anyone wishing to translate machine learning services. Furthermore, the machine learning ecosystem is no different than the programming ecosystem at large when it comes to the predominance of English. A significant portion of all publicly available data and models are in English, while the rest of the human languages are under-represented.

This has resulted in a sort of vacuum when it comes to speech recognition models. While state-of-the-art models for English speech recognition are being created, constantly improving upon previous research, there is significantly less research done on the rest of the world language. These state-of-the-art models are often created by researchers from well-funded organizations or companies, with the resources to create models with very big memory and storage footprints and to train them with large amounts of data. When researchers wish to build models for less common languages, the scarcity of data, models and computing resources can be daunting to handle.

This work proposes to solve these three problems by using a model with a small memory and storage footprint designed to recognize individual phonemes. Phonemes are the smallest vocal components of speech that distinguish between words and are considered one of the basic building blocks of human languages. While languages can have hundreds of thousands of words, these can typically be broken down into several dozens of distinct phonemes at most. English for example is estimated to contain around 171,000 words (Sagar-Fenton and McNeill, 2018), but only 42 phonemes.

While it has been argued (Garling, 2015) that phonemes might be an artificial human construct, making them a non-optimal way to train machine learning models, this is just as true as stating that any written symbol such as a word, a letter or a numerical token is an artificial human construct. The fact stands however that predictive machine learning models map between input data and target symbols, making phonemes a valid choice as output symbols.

The same sources claim it is better to train such models to find a direct mapping between input data and the raw predicted text. However, choosing phonemes as our output symbols has two major advantages over direct mapping to output words:

First, because the number of distinct phonemes in any human language is several orders of magnitude smaller than the number of words (or sentences) in that language, we can train the model on much less training data. Even if our training data consists of a single utterance of every word in a language, it will inadvertently encompass all the phonemes in that language a hundred times over.

Second, the common way to represent output symbols in machine learning systems is by representing them with vectors. Because of the sheer number of distinct words in any given language, encoding them as vectors requires embedding them in a very high dimensional vector space. Even using methods such as word embeddings would still result in a vector space with a much bigger dimension than any space used to embed phonemes. The most common way to map from internal model representation to output vectors is by using fully connected (dense) layers, which perform a matrix multiplication as part of their calculations. Since naïve matrix multiplication has a time complexity of $O(n^3)$, even small increases to the size of the output vector space can cause a significant increase in training times, as well as in noticeable delays in inference time when deploying the model. The small number of distinct phonemes in human languages mitigates this problem.

Experimentally, this work demonstrates that phoneme recognition models made using the latest research can achieve performance comparable to current state of the art models, while only using a fraction of the data required for distinct word recognition. Such models can be created and trained very quickly, and later be deployed to an edge device for real time speech recognition.

Log-Mel spectrograms:

One of the most studied transforms in mathematics is the discrete Fourier transform (DFT), which allows us to convert equally spaced samples of a function to matching samples of the function's discrete-time Fourier transform (DTFT). The DFT allows us to convert discrete samples measuring the amplitude of human voice to samples measuring the frequency components of the same voice.

When analyzing recordings of human voice, it is often desirable not to convert the entire record at once, but to convert it in separate consecutive frames. If a recording contains multiple phonemes, the segmenting process means the frequencies for each phoneme are calculated separately. This allows us to train machine learning models to classify the phonemes accordingly. The practice of segmenting the DFT into frames in this manner is known as the short-time Fourier transform (STFT).

The converted and transformed frames can be displayed in what we call a spectrogram. In the simplest form, the frames are stacked side-by-side with the frequencies in each frame stacked vertically. This process results in a grid-like image where each grid cell represents the amplitude of a given frequency, within a specific frame. The cells are then colored according to their amplitude, resulting in a visual representation of the original signal.

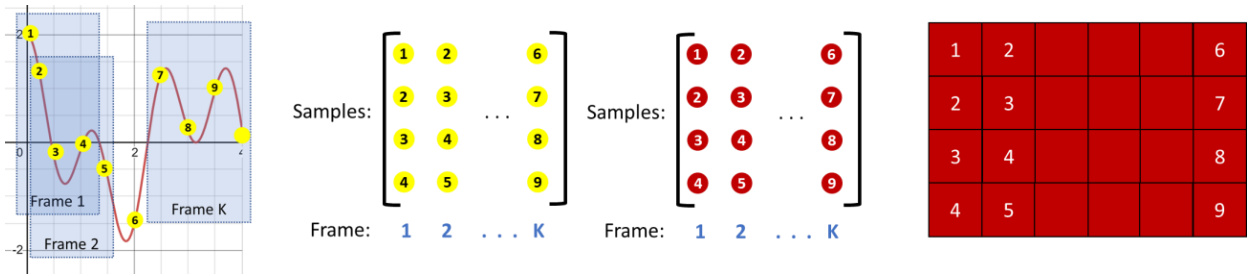

**Figure 1:** Conversion from audio to spectrogram in four steps. Left to right: 1) The raw audio is sampled at regular intervals. 2) the discrete samples are segmented into frames and stacked side-by-side. 3) We apply the DFT to each frame to obtain the matching samples on the DTFT of the original audio. 4) When displayed as an image, the DTFT matrix shows us the spectrogram.

Converting the voice signals into images allows us to transfer the problem of speech recognition from the audio domain to the image domain. This in turn allows us to apply powerful tools used for image processing to aid us in speech recognition. The process has been well demonstrated using a convolutional neural network (CNN) to classify audio signals using their spectrograms (Arik and Kliegl et al., 2017).

While modern microphones record sound record sound using tens of thousands of samples every second, previous studies have shown us that human voice is distinguished at much lower frequencies (Baken, 1987). Due to this fact, raw spectrograms are less meaningful than they could be. One way to ameliorate this is by measuring the spectrogram frequencies in decibels, instead of the raw amplitudes. Since decibels progress on a logarithmic scale, the wide range of frequencies present in spectrograms is displayed in a more manageable scale.

Amplitude is normally measured in Hertz (Hz), the standard measure unit for frequency. However, when comparing equidistant sounds in Hz, humans do not perceive them to be the same distance from one another. One alternative measurement scale is the Mel scale (mel), measuring the human perceived pitch rather than the true pitch (Stevens, 1937). Spectrograms measuring frequencies in mel scale allocate more frequency bands to the lower frequencies in which human languages are spoken, therefore providing more data for training speech recognition models.

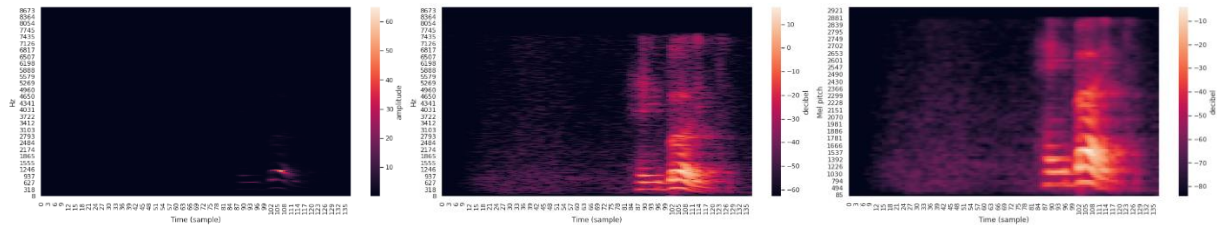

**Figure 2:** A two second audio sample converted into spectrograms. The left spectrogram is measured using Hz scale, showing the amplitude of each frequency across time. The middle spectrogram is identical to the left, with frequencies converted to decibels. The right spectrogram is again identical, with frequencies measured in Mel scale and converted to decibels.

Spectrograms in which the amplitude is converted from Hertz scale to Mel scale, and then converted to decibels are known as Log-Mel spectrograms.

Convolutional recurrent neural networks:

Convolution kernels are famous for being invariant to the location of the features they are trained to detect. Phoneme ordering, on the other hand, is not; Changing the order of the recognized phonemes also changes the meaning of the output.

A design pattern that is becoming more common in recent years (Zuo et al., 2015) uses the order-dependent nature of recurrent neural networks to learn not only the right output tokens, but also their correct order, by first passing them through a convolution layer and then through a recurrent layer. When applied to phoneme recognition, this allows the convolution layer to learn the necessary activations for detecting the phonemes present in the speech while the recurrent layer learns their position relative to each other.

Sequence to sequence models:

Recurrent neural networks (RNNs) in their standard form (Jordan, 1986) receive two separate inputs: an input vector and a "hidden" state vector. The output is also made of two vectors: an output vector that matches the input vector, and an updated hidden state vector. The updated hidden state vector is used to override the previous hidden state, and is fed to the model along with the next input vector.

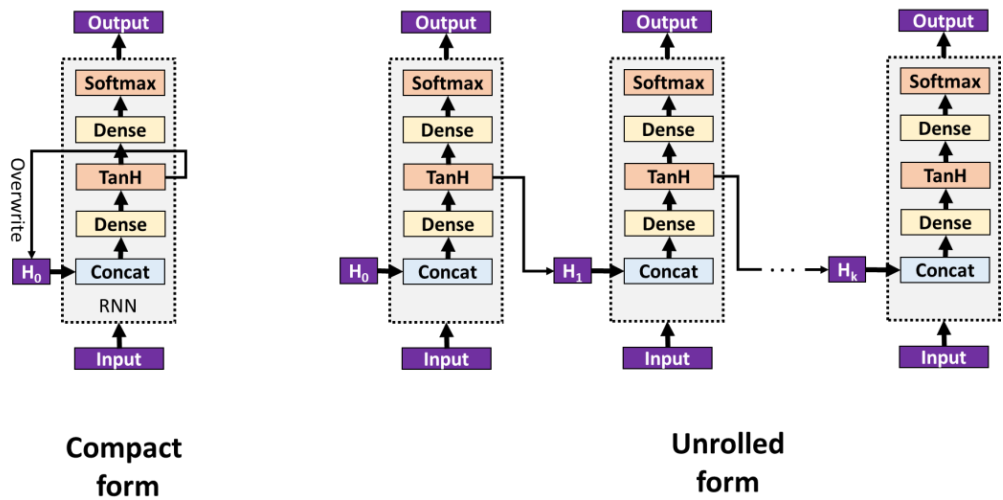

**Figure 3:** The structure of a standard recurrent neural network, displayed in two ways. The left illustration shows the actual structure of the RNN, with arrows indicating the connections between the computational steps. The right illustration shows the "unrolled" form, which better explains how sequences of inputs are treated by the network. "$H_0$", "$H_1$" ... are used to denote the hidden state vector.

In sequence-to-sequence models (seq2seq in short) we use not one but two recurrent neural networks to predict or generate output sequences. Pioneered by a research team at Google (Sutskever et al., 2014), seq2seq models use two recurrent neural networks called the "encoder" and the "decoder". First, the sequence of inputs is fed one by one to the encoder network, which updates its hidden state vector once for every input. Then, the encoder passes its hidden state vector to the decoder, which uses it as its own hidden state vector. The premise behind sharing the hidden state vector is that the encoder embeds the entire sequence of inputs as a single vector in a lower dimensional space, which the decoder then uses to output a sequence representing the original input.

Once we pass the hidden state vector from the encoder to the decoder, we discard the encoder outputs and input an initialization vector to the decoder. Every output vector is recorded and might be further processed or just converted to the token it represents. The model has a predefined vector called the "end-of-sequence" vector. Whenever the decoder outputs this vector (or a predetermined counter has been reached), we exit the loop and evaluate the sequence of output vectors produced by the decoder.

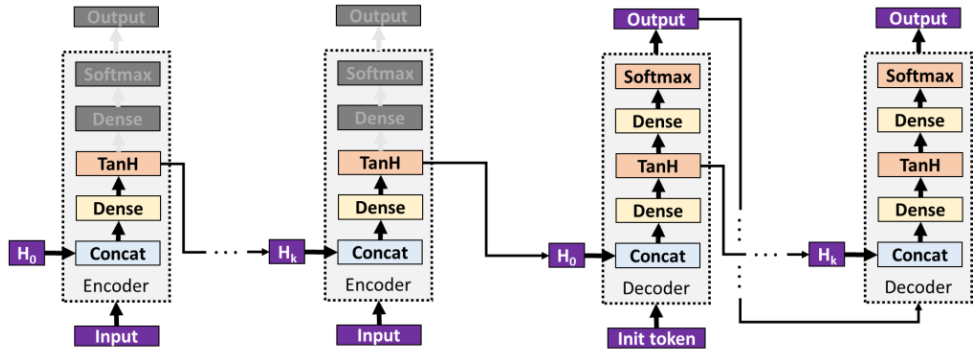

**Figure 4:** The structure of a sequence-to-sequence model, shown in unrolled form. The encoder outputs are discarded, and the final hidden state is used to initialize the decoder. The first decoder input is just a predefined "initialization token", represented as a vector. Each decoder output is fed back as the input of the next step, along with the curried hidden state vector.

This is the place to note there are several improvements over the vanilla RNN model, with the two most popular ones being long-short term memory networks (LSTMs) and gated recurrent units (GRUs). While LSTM network have been known for over two decades (Hochreiter and Schmidhuber, 1997), GRUs are a more recent improvement (Cho et al., 2014). In a recent study[12] (Chung et al., 2014) the three designs have been compared to each other. Both the LSTM and the GRU outperformed the vanilla RNN design, while achieving comparable results to one another. Since GRUs have a lower parameter count than LSTMs, this makes them better suited for deployment on low-resource edge devices. In addition, the lower parameter count also makes forward and backward passes through GRUs faster than through LSTMs.

Attention mechanism:

Working with long sequences is not easy for machine learning models, with accuracy tending to decrease the bigger the length of the input and output sequences. Even humans would struggle to handle long sequences in the same fashion - imagine translating a paragraph of 100 words which you have only read once, without being allowed to look back at words you have already read. The hidden state of the model gets overwritten again and again, reducing the ability of initial inputs to influence the later outputs in the sequence.

Attention mechanisms (Bahdanau et al., 2014) seek to correct this by weighting in all the encoder outputs, instead of just the last hidden state. Instead of discarding them as we do in the vanilla RNN, the decoder network assigns relevant weights to each encoder output at every iteration of the decoder loop before passing them through a softmax layer (Bridle, 1989). These weights are called the "alignment vector" of the attention mechanism. In one of the more popular variants of the attention mechanism (Luong et al., 2015), we then multiply the alignment vector and the encoder outputs to produce a "context vector", containing the dot product of the alignment vector with each encoder output. This determines "How much attention" should be paid to each encoder output.

Once we have calculated the alignment and context vectors, we concatenate the raw decoder outputs with the context vector. The concatenated vector is passed through a fully connected layer (with a nonlinear activation function) to generate the output vector for the current iteration of the decoder loop.

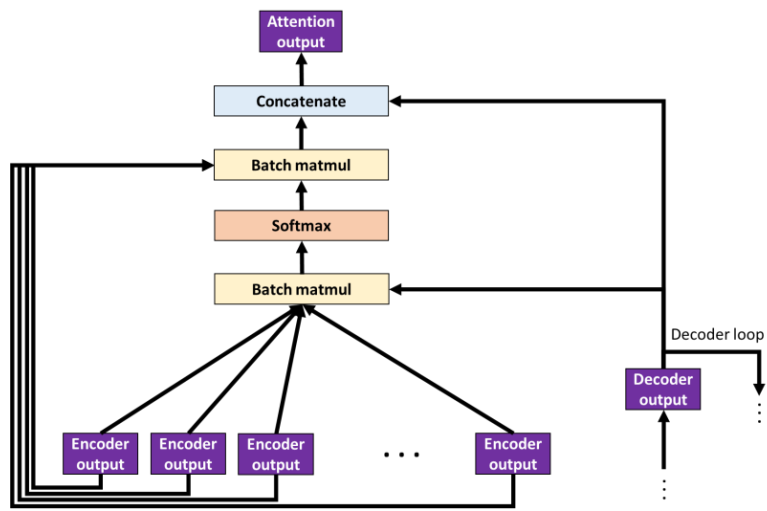

**Figure 5:** An attention mechanism based on the Luong variant, using the dot product operation (implemented as batch matrix multiplication). The decoder output is used as a weight vector to determine the significance of each encoder output, yielding the alignment vector as the output of the softmax layer. The context vector is the output of the 2$^{nd}$ batch matrix multiplication layer.

Recording and pre-processing:

The end-to-end task used to train the model and deploy it involves voice activation of an air conditioner using speech commands in Hebrew. First, around 45 minutes of audio were recorded, resulting in 17 distinct phrases: 14 variations of 4 distinct commands: "Turn on", "Turn off", "Heat the room" and "Cool the room", and 3 additional phrases representing irrelevant sentences in Hebrew ("Grumpy old man", "Bring the shampoo", "There's an oak here"). The phonemes in these phrases partially overlap the phonemes in the 14 meaningful phrases, with the relative complement of phonemes tagged as "unknown phonemes".

In addition, approximately 71 minutes of non-verbal background sounds were sampled from sources available under public license of fair-use license. These were used to train the model to distinguish between speech presence and absence.

Recording the speech commands was done by a single speaker in a silent environment, resulting in around ~1310 utterances. Each utterance was then padded to 2 seconds in length to ensure they have the same sample count. The background sound recordings were sliced into 2-second-long clips. These resulted in around ~2140 clips of background noise for a total of 3450 audio clips.

The commands consisted of 26 spoken phonemes, which were mapped to integer tokens representing them. An additional five tokens were added, representing "unknown phoneme" (UNK), "begin speech" (BEG), "end speech" (END), "silence" (BKG) and "non-verbal noise" (NOISE). These add up together to form 31 tokens.

All the audio files were recorded in 16-bit PCM WAV format, and down-sampled to a non-standard rate of 8820hz. This unusual choice has several reasons:

- The lower the sample rate the faster the real-time conversion of the user's speech to log-mel spectrograms. The conversion requires two computationally intense operations: A Fourier transform, which can be calculated using the Cooley-Tukey algorithm (Cooley and Tukey, 1965) at $O(n * \log(n))$ time complexity, and a Matrix multiplication at $O(n^3)$ time complexity.
- The target edge device used in the experiments was running the Android operating system, which only guarantees recording audio at a sample rate of 44100 Hz. While audio can be converted to any frequency by naïve sampling, high quality down-sample algorithms require an up-sample step using interpolation and a low-pass filter, before performing a down-sample step with another low-pass filter. The up-sample step can be avoided if the ratio between the original sample rate and the target sample rate is a whole number (i.e., not a fraction). 8820 Hz is an integer factor of 44100Hz, thus fulfilling this requirement.

- The lower the sample rate, the less audible the recordings are. Lower quality audio unavoidably results in lower accuracy rates for the model. While we want to reduce the amount of time it takes to recognize syllables, we want the results to remain accurate. The choice of 8820 Hz was due to the high levels of recognition accuracy at this sample rate, while keeping conversion times on the (admittedly old) edge device shorter than 1 second.

The audio clips were transformed into spectrograms using a frame length of 1024 samples and a frame overlap of 128 samples. They were then converted to mel scale using 80 discrete bins for the frequency range between 0Hz and 4410Hz.

When creating spectrograms from user speech on the edge device, a performance gain was achieved if the spectrograms were generated with a 90 degrees rotation relative to the "standard" orientation of a spectrogram. Therefore, all spectrograms used for training were rotated 90 degrees clockwise before being fed to the model as input.

Model:

The model is composed of four sections: two convolution blocks, two stacked seq2seq layers, one attention block and two fully connected blocks. The total parameter count for the model is ~307k parameters[1].

The spectrograms are first fed to the convolution blocks. Each convolutional block is composed of three layers: a convolution layer, an activation layer and a normalization layer. Both convolution layers use 5x1 filters and a stride of 1. They also zero-pad their inputs by two pixels vertically (2x0 zero padding). The first convolution layer has 10 filters while the second convolution layer has just one. Both convolution blocks use Leaky Rectified Linear Units (Leaky RelU) as their nonlinear activation functions (Dahl et al., 2013), and Batch Normalization (Ioffe and Szegedy, 2015) in the normalization layer during training time.

The encoder block is made of two stacked gated recurrent units (GRUs), each containing two GRU cells forming a bidirectional GRU (Cho et al., 2014). The output of the convolutional layers is then viewed as a single-channel image and split into columns. The columns are treated as discrete time samples and fed in sequence to the first forward-facing GRU cell, and again in reverse sequence to the backward-facing GRU cell. Each of the two GRU cells has its own hidden state vector, and the outputs are passed as inputs to the matching GRU cell in the second bidirectional GRU.

---

[1] The model and code described in this paper are publicly available at
https://github.com/yonatankarimish/YonaVox

The decoder block is also made of two stacked GRUs. However, since we do not know the output vectors during real time inference we cannot use bidirectional GRUs. Each GRU thus contains a single GRU cell, which is initialized by concatenating the hidden state vectors of the respective bidirectional GRU cells in the encoder. The encoder uses four hidden state vectors of shape 1x64, which form two 1x128 vectors once concatenated for the decoder.

Both encoder and decoder outputs are passed through an attention block, as described in the section above regarding the attention mechanism. The attention block passes the concatenated context + decoder vector to two fully connected (dense) blocks in succession. The first dense layer has a 256x64 weight matrix and a 1x64 bias matrix. The second dense layer has a 64xT weight matrix and a 1xT bias vector, where T equals the number of distinct tokens (31 in our case). Both dense blocks use a hyperbolic tangent (TanH) as their nonlinear activation functions (Haykin, 1998), finally outputting a logit vector with a shape of 1xT.

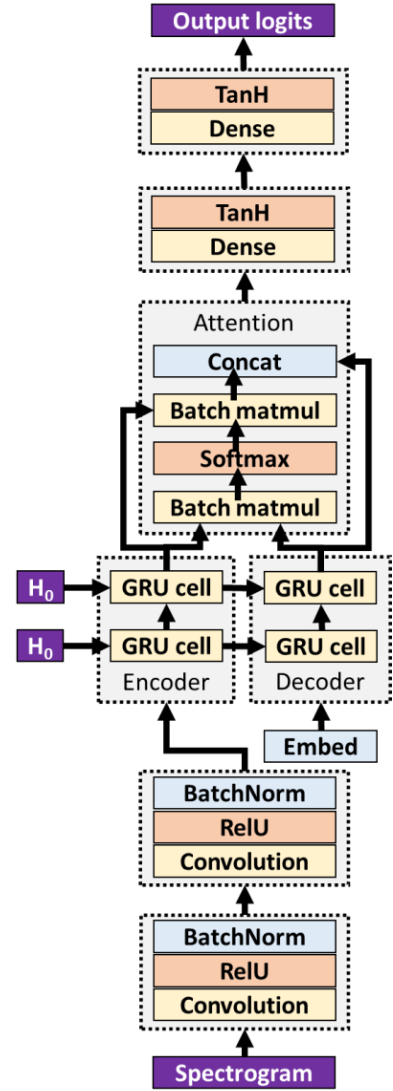

**Figure 6:** Architecture of the complete model, made of two convolution blocks, an encoder-decoder, an attention block and two dense blocks.

The design of the convolution blocks and the seq2seq layers was partially influenced by the works of de Andrade et al., 2018.

Loss function:

The standard loss function used for sequence models is the categorical cross-entropy, defined as $\mathcal{L}_{cce}(\hat{y},\ y) = -\hat{y} \otimes \ln(\sigma(y))$

Where:

- $\hat{y}$ = matrix of one-hot encoded ground truth tokens.
- $y$ = matrix output of model, where adjacent columns represent model output as consecutive iterations of the decoder loop.
- $\sigma$ = Softmax activation function, calculated independently for each column of y.
- $ln$ = natural logarithm function, calculated independently for each scalar of $\sigma(y)$.
- $\otimes$ = Element-wise product (Hadamard product) between $\hat{y}$ and $\ln(\sigma(y))$.

However, during initial training of the model, it became evident that the END, BKG and NOISE tokens were over-represented in the data due to padding the audio clips themselves and due to padding the token sequences to equal length for training efficiency. This hampered the ability of the model to learn the correct phonemes present in the input spectrograms and gave a non-representative metric for the model's accuracy.

To fix this, we calculate the inverse of the occurrence frequency for each phoneme present in the training set. We then create two vectors: a weight vector $w$ where $w_i$ is defined as the inverse frequency of $token_i$, and an occurrence vector $f$ where $f_i$ is defined as the number of times $token_i$ is present within the list of training set phonemes.

Once we calculate $w$, we can modify our loss function to the following loss:

$$\mathcal{L}_{wcce}(\hat{y}, y) = -\frac{S}{W} \sum_{i=1}^{H} \sum_{j=1}^{W} w_j * \hat{y}_{ij} * \ln(\sigma(y_{ij}))$$

Where:

- $H, W$ = height and width of both $\hat{y}$ and $y$ ($W = T$).
- $S$ = loss scale coefficient, defined as $S = \frac{1}{T} * \sum_{i=1}^{T} f_i$ and representing the mean occurrence of any token across the entire training set.

Training:

Quite a few training configurations were attempted, which involved many changes to the model hyper parameters and to the training data itself. During these experiments it became obvious that tweaking the hyper parameters had little effect on the final accuracy levels of the model, while modifying the data had the most effect. The best performing models were trained on non-augmented training data of the original recordings, as collected in the "recording and pre-processing" section above.

When implementing the decoder loop, the model would occasionally replace the decoder predictions from the previous loop iteration with the ground truth tokens, with a random probability of 50%. This practice is known as teacher forcing (Williams and Zipser, 1989) and is used to prevent errors from propagating during the forward pass when generating outputs, or during the backward pass when calculating gradients.

The data was split into a training set of 3104 spectrograms and a test set of 346 spectrograms. Each dataset had samples of all the different commands and noise types. The model was then trained with a batch size of 1024 spectrograms for 1200 epochs. Training was done on an Nvidia Tesla P100 GPU and took around ~35 minutes to complete.

Model loss was calculated using the weighted categorical cross-entropy function described above, and back-propagation was done using the Adam algorithm (Kingma and Ba, 2015) with a learning rate of $\alpha = 0.001$ and hyper-parameters $\beta_1 = 0.9, \beta_2 = 0.999$. All code used for training the model was written in Python, using the popular NumPy(Harris and Millman, 2020) and PyTorch (Paszke et al., 2019) libraries.

Deployment:

Once the training was done, the model was deployed on a mobile phone running an Android 7 operating system with 1.5GB RAM. The model itself occupies only 1238kb of storage while the interpreter uses an additional 30.9mb of storage.

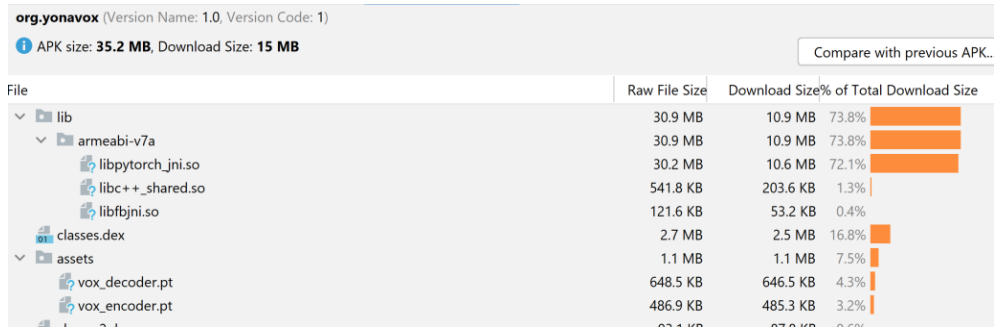

**Figure 7:** Memory footprint of the mobile app, deployed to the edge device. Most of the space is occupied by the model interpreter, which is written as native code in C++. The recognition model and accompanying code constitute less than 13% of the app size when installed.

Each deployed model was compared to the latest trained model by forwarding them both an empty spectrogram (i.e., a spectrogram consisting of zeros). Both outputs were verified to ensure the correct model was deployed on the edge device.

The android app was then connected to an IoT sensor through a web server exposing an API. When a user speaks near the mobile phone, the speech is down sampled, converted to a log-mel spectrogram and forwarded as input to the model. Once the model outputted its predictions the sensor was activated using HTTP requests to the correct API endpoint. The total inference time for live speech data was around ~1810ms, of which the model forward pass was around ~450ms. Occasional deployments to newer devices consistently resulted in shorter inference times.

Results:

The model displayed good performance throughout its entire training process. To evaluate the performance of the model, a weighted accuracy metric was developed for the same reasons a weighted loss function was used during training.

For the weighted accuracy metric, we consider two token vectors to be equal if they are both of equal length, and if they both contain the same tokens in the same order. When comparing the token vectors, every predicted token that matches the true token is multiplied by its weight from the inverse weight vector $w$, and all the true tokens are multiplied by their respective weights as well. The true tokens and the correctly predicted tokens are separately summed, and we then divide the sum of correctly predicted tokens from the sum of the true tokens.

The above metric can be formulated as

$$\mathcal{M}_{wacc} = \frac{\langle w^{r\&p}, r\&p \rangle}{\langle w^r, r \rangle}$$

Where:

- $\langle x, y \rangle$ denotes the standard inner product in $\mathbb{R}^T$ for every two arbitrary vectors $x, y \in \mathbb{R}^T$
- $\&$ denotes the element-wise equality comparison for every two arbitrary vectors $x, y \in \mathbb{R}^T$, and is defined as: $[x \& y]_i = \begin{cases} 1 & if\ x_i = y_i \\ 0 & otherwise \end{cases}$
- $r$ = vector containing the ground-truth tokens for a given audio file
- $p$ = vector containing model predictions for a given audio file
- $w^r$ is a vector in which $w_i^r = w_{r_i}$
- $w^{r\&p}$ is similarly defined as $w_i^{r\&p} = \begin{cases} w_{p_i} & if\ r_i = p_i \\ 0 & otherwise \end{cases}$

After the 1200 epochs used for training, the weighted accuracy was calculated over the test set of the model. The average weighted accuracy was 93.21%.

Another common metric used for evaluating speech recognition systems is called the "phoneme error rate" (PER), which can be viewed as an extension of a non-weighted accuracy metric that also considers token insertions and deletions. To calculate the PER between two token vectors (Thoma, 2013), we first compute the Levenshtein distance between the two vectors, then divide the distance by the length of the ground truth vector. The average PER calculated for the test set of the model was 1.77%.

To visualize the model's performance on the test set, a confusion matrix (Townsend, 1971) was calculated for all the phonemes present in the speech data. A confusion matrix allocates a row for every distinct ground truth token, and a column for every distinct predicted token. Whenever a token is predicted, the cell intersecting between the predicted value and the actual value increases in value, resulting in a matrix which can be color coded to visualize the model accuracy.

A perfect model will result in an identity matrix, while random predictions will result in matrices where all cells converge towards an equal value.

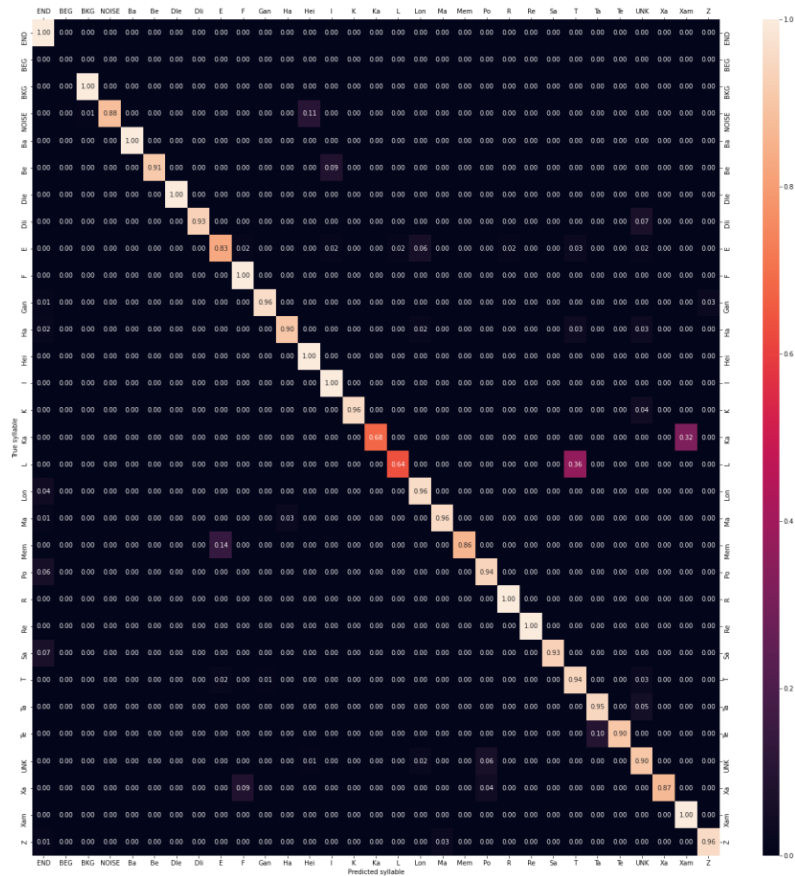

**Figure 8:** Confusion matrix for the trained model, calculated over the test set. The high accuracy rate of the model reflects in the matrix by the strong color contrast between the main diagonal and the rest of the cells. Two noticeable errors can be seen to the right of the matrix: "L" -> "T" and "Ka" -> "Xam".

Discussion:

The biggest drawback of the training process would probably be using speech data from a single speaker only. It can be assumed that metric scores for training on speech data originating from multiple speakers would be lower than the ones presented above. However, training the model for a larger number of epochs and the increase in sample counts from the additional speakers would probably mitigate some (if not most) of the decrease in model performance.

Having said that, one should keep in mind that the latest speech recognition models are much larger than this model and train on significantly larger quantities of data. The latest state-of-the-art model at the time of writing this paper for English speech recognition, called Wav2Vec2.0 (Baevski et al., 2020) has 317 million parameters compared to the 307k parameters of this model. Yet in order to achieve equivalent metric scores (a Word error rate of 1.6%, in their case) they had to train on 100 hours of the labeled LibriVox 60k dataset. That's more than 1000 times the model size and more than 125 times the total amount of speech data.

While these scores are not truly comparable to one another, as they were obtained while training on different datasets, they still show that this model performs quite well. Phoneme error rates of 1.77% and weighted accuracy of 93.21% are not only comparable to many of the latest models, but also allow the model to be deployed in production applications and services.

The low disk space the model occupies, and its small memory requirements can potentially allow for real-time prediction on even the smallest of edge devices. The biggest constraint to deployment itself is surprisingly not the model size but rather the size of the interpreter. The mobile app is deployable on most mobile phones but would probably be restricted from deploying to devices such as microprocessors and IoT sensors. There exists a process for reducing the interpreter size by generating a custom build from its own source code using only the subset of methods required by the model itself. Doing so would probably reduce the disk space it occupies by a significant percentage, making such deployments more feasible.

When running real-time speech recognition, the model consistently succeeds in understanding the user and controlling the air conditioner sensor based on his spoken commands. However, the model has varying degrees of success when transcribing speech from users other than the speaker used for recording the training data. This would probably be fixed if more speakers would contribute to the speech data, as mentioned above when discussing the model drawbacks. For demonstration purpose, successfully transcribing speech from a similar distribution as the training set the model was trained on is more than enough, and the model does exactly that.

Conclusions:

Rather than create yet another model that tries to break a state-of-the-art record on one of the common datasets, this work proves that small models can be trained using little data to very high levels of accuracy and consistently recognize phonemes with very low error rates.

This work also proves these models can be deployed to many types of edge devices such as mobile phones, tablets, and car displays - to name a few. Smart design choices allow for fast real time speech recognition on these devices, which can be used for any number of purposes.

These findings will hopefully encourage other researchers to create similar models in their own native languages, which will ultimately make speech recognition more accessible to people throughout the world.

References:

Arik, S. O., Kliegl, M., Child, R., Hestness, J., Gibiansky, A., Fougner, C., Prenger, R., & Coates, A. (2017). Convolutional Recurrent Neural Networks for Small-Footprint Keyword Spotting. ArXiv:1703.05390 [Cs]. http://arxiv.org/abs/1703.05390

Baevski, A., Zhou, H., Mohamed, A., & Auli, M. (2020). wav2vec 2.0: A Framework for Self-Supervised Learning of Speech Representations. ArXiv:2006.11477 *[Cs, Eess]*. http://arxiv.org/abs/2006.11477

Bahdanau, D., Cho, K., & Bengio, Y. (2016). Neural Machine Translation by Jointly Learning to Align and Translate. *ArXiv:1409.0473 [Cs, Stat]*. http://arxiv.org/abs/1409.0473

Baken, R. J., & Orlikoff, R. F. (2000). *Clinical measurement of speech and voice* (2nd ed). Singular Thomson Learning.

Bridle, J. S. (1990). Probabilistic Interpretation of Feedforward Classification Network Outputs, with Relationships to Statistical Pattern Recognition. In F. F. Soulié & J. Hérault (Eds.), *Neurocomputing* (pp. 227–236). Springer Berlin Heidelberg. https://doi.org/10.1007/978-3-642-76153-9_28

Cho, K., van Merrienboer, B., Gulcehre, C., Bahdanau, D., Bougares, F., Schwenk, H., & Bengio, Y. (2014). Learning Phrase Representations using RNN Encoder-Decoder for Statistical Machine Translation. *ArXiv:1406.1078 [Cs, Stat]*. http://arxiv.org/abs/1406.1078

Chung, J., Gulcehre, C., Cho, K., & Bengio, Y. (2014). Empirical Evaluation of Gated Recurrent Neural Networks on Sequence Modeling. *ArXiv:1412.3555 [Cs]*. http://arxiv.org/abs/1412.3555

Cooley, J. W., & Tukey, J. W. (n.d.). *An Algorithm for the Machine Calculation of*

*Complex Fourier Series*. 5.

Crystal, D. (2003). *English as a global language* (2nd ed). Cambridge University Press. https://www.google.co.uk/books/edition/_/d6jPAKxTHRYC?hl=en&gbpv=1&pg=PP1

Dahl, G. E., Sainath, T. N., & Hinton, G. E. (2013). Improving deep neural networks for LVCSR using rectified linear units and dropout. *2013 IEEE International Conference on Acoustics, Speech and Signal Processing*, 8609–8613. https://doi.org/10.1109/ICASSP.2013.6639346

de Andrade, D. C., Leo, S., Viana, M. L. D. S., & Bernkopf, C. (2018). A neural attention model for speech command recognition. *ArXiv:1808.08929 [Cs, Eess]*. http://arxiv.org/abs/1808.08929

Garling, C. (2015, February 2). *Google Brain's Co-inventor Tells Why He's Building Chinese Neural Networks | by Caleb Garling | Backchannel | Medium*. Medium.Com. https://medium.com/backchannel/google-brains-co-inventor-tells-why-hes-building-chinese-neural-networks-662d03a8b548

Harris, C. R., Millman, K. J., van der Walt, S. J., Gommers, R., Virtanen, P., Cournapeau, D., Wieser, E., Taylor, J., Berg, S., Smith, N. J., Kern, R., Picus, M., Hoyer, S., van Kerkwijk, M. H., Brett, M., Haldane, A., del Río, J. F., Wiebe, M., Peterson, P., … Oliphant, T. E. (2020). Array programming with NumPy. *Nature*, *585*(7825), 357–362. https://doi.org/10.1038/s41586-020-2649-2

Haykin, S. S. (1999). *Neural networks: A comprehensive foundation* (2nd ed). Prentice Hall.

Hochreiter, S., & Schmidhuber, J. (1997). Long Short-Term Memory. *Neural Computation*, *9*(8), 1735–1780. https://doi.org/10.1162/neco.1997.9.8.1735

Ioffe, S., & Szegedy, C. (2015). Batch Normalization: Accelerating Deep Network Training by Reducing Internal Covariate Shift. *ArXiv:1502.03167 [Cs]*. http://arxiv.org/abs/1502.03167

Jordan, M. I. (1986). *Serial order: A parallel distributed processing approach*. Institute for Congitive Science, University of California, San Diego. https://cseweb.ucsd.edu/~gary/258/jordan-tr.pdf

Kingma, D. P., & Ba, J. (2017). Adam: A Method for Stochastic Optimization. *ArXiv:1412.6980 [Cs]*. http://arxiv.org/abs/1412.6980

Luong, M.-T., Pham, H., & Manning, C. D. (2015). Effective Approaches to Attention-based Neural Machine Translation. *ArXiv:1508.04025 [Cs]*. http://arxiv.org/abs/1508.04025

Paszke, A., Gross, S., Massa, F., Lerer, A., Bradbury, J., Chanan, G., Killeen, T., Lin, Z., Gimelshein, N., Antiga, L., Desmaison, A., Köpf, A., Yang, E., DeVito, Z., Raison, M., Tejani, A., Chilamkurthy, S., Steiner, B., Fang, L., … Chintala, S. (2019). PyTorch: An Imperative Style, High-Performance Deep Learning Library. *ArXiv:1912.01703 [Cs, Stat]*. http://arxiv.org/abs/1912.01703

Sagar-Fenton, B., & McNeill, L. (2018, June 24). *How many words do you need to speak a language? - BBC News*. Bbc.Com. https://www.bbc.com/news/world-44569277

Stevens, S. S., Volkmann, J., & Newman, E. B. (1937). A Scale for the Measurement of the Psychological Magnitude Pitch. *The Journal of the Acoustical Society of America*, *8*(3), 185–190. https://doi.org/10.1121/1.1915893

Sutskever, I., Vinyals, O., & Le, Q. V. (2014). Sequence to Sequence Learning with Neural Networks. *ArXiv:1409.3215 [Cs]*. http://arxiv.org/abs/1409.3215

Thoma, M. (2013, November 15). *Word Error Rate Calculation*. Martin Thoma. https://martin-thoma.com/word-error-rate-calculation/

Townsend, J. T. (1971). *Theoretical analysis of an alphabetic confusion matrix. Perception & Psychophysics*. https://psymodel.sitehost.iu.edu/publications_all.shtml. https://psymodel.sitehost.iu.edu/papers/tow71theoretical.pdf

Williams, R. J., & Zipser, D. (1989). A Learning Algorithm for Continually Running Fully Recurrent Neural Networks. *Neural Computation*, *1*(2), 270–280. https://doi.org/10.1162/neco.1989.1.2.270

Zuo, Z., Shuai, B., Wang, G., Liu, X., Wang, X., Wang, B., & Chen, Y. (2015). Convolutional recurrent neural networks: Learning spatial dependencies for image representation. *2015 IEEE Conference on Computer Vision and Pattern Recognition Workshops (CVPRW)*, 18–26. https://doi.org/10.1109/CVPRW.2015.7301268